\documentclass[final]{cvpr}

\usepackage{times}
\usepackage{epsfig}
\usepackage{graphicx}
\usepackage{amsmath}
\usepackage{amssymb}

\usepackage{enumitem}
\usepackage{float}
\usepackage[font=small,skip=-1pt]{caption}

\usepackage[pagebackref=true,breaklinks=true,colorlinks,bookmarks=false]{hyperref}

\def\cvprPaperID{14} 
\def\confYear{CVPR 2021}

\graphicspath{{images/}}

\begin{document}

\title{Reconsidering CO2 emissions from Computer Vision}

\author{Andre Fu\\
University of Toronto\\
{\tt\small andre.fu@mail.utoronto.ca}
\and
Mahdi S. Hosseini\\
University of New Brunswick\\
{\tt\small mahdi.hosseini@unb.ca}
\and 
Konstantinos N. Plataniotis\\
University of Toronto\\
{\tt\small kostas@ece.utoronto.ca}
}
\maketitle

\begin{abstract}
    Climate change is a pressing issue that is currently affecting and will affect every part of our lives. It's becoming incredibly vital we, as a society, address the climate crisis as a universal effort, including those in the Computer Vision (CV) community. In this work, we analyze the total cost of CO2 emissions by breaking it into (1) the architecture creation cost and (2) the life-time evaluation cost. We show that over time, these costs are non-negligible and are having a direct impact on our future. Importantly, we conduct an ethical analysis of how the CV-community is unintentionally overlooking its own ethical AI principles by emitting this level of CO2. To address these concerns, we propose adding ``enforcement'' as a pillar of ethical AI and provide some recommendations for how architecture designers and broader CV community can curb the climate crisis.
\end{abstract}

\section{Introduction}
Our Society is experiencing an exponential growth in developing, training, and using deep-learning pipelines, particularly within Computer Vision (CV), the rapid growth has even surpassed human-level cognition \cite{wu2015deep}. Fueled by advances in computing, the state-of-the-art (SOTA) CV models are consistently beating benchmarks, with the most computationally intensive models achieving SOTA \cite{lu2021neural}. Consequently, developing and training SOTA models demand an inordinate computational burden, which corresponds to energy and environmental costs. Currently, there exist two methods for developing CV models, (1) Heuristic hand-crafted model development; and (2) CV Neural Architecture Search (CV-NAS) algorithms. 


In order to design SOTA models, using either hand-crafted or NAS methods, some form of searching for an optimal architecture is required. With hand-crafted networks, the search phase is the iterative training for optimization and with NAS, the search is embedded directly into the architecture generation pipeline. After architectures get frozen they are released to the public which leads to wide-scale adoption of popular architectures such as ResNet \cite{he2016deep}, VGG \cite{simonyan2014very}, and GoogLeNet \cite{szegedy2015going} just to name a few. Then models experience an evaluation phase, over the life-time of these architectures, they are trained and evaluated a great number of times. Both these phases require enormous computational costs and equivalently enormous CO2 emissions. As such, an important area to consider is the CO2 impacts of the CV community, the ethical considerations of those impacts, and how we can curb the global climate crisis.

\begin{figure}[t!]
\begin{center}
\includegraphics[width=1.0\linewidth]{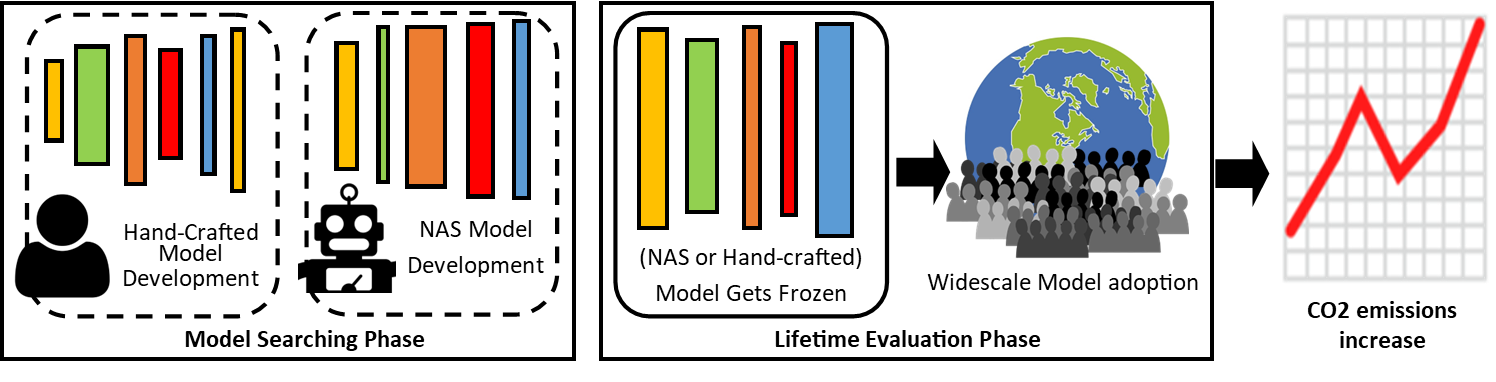}
\end{center}
\caption{The two phases of CO2 emissions: model development through search and life-time evaluation. As a model (with high computational footprint i.e. FLOPS) gets world-wide adoption, the CO2 emissions increase.}
\label{fig:Concept_Figure}
\vspace{-1mm}
\end{figure}

Within CV no such analysis of the CO2 has been conducted, and therefore this paper aims to do so; guided by the following questions:
\begin{itemize}
    \item During the model-searching phase, how much CO2 is being emitted, and how can we curb these emissions?
    \vspace{-0.3cm}
    \item Over the lifetime of a model, can we quantify the total amount of CO2 emitted and recommend solutions to curb these emissions? 
    \vspace{-0.3cm}
    \item Who are the marginalized communities affected by CO2 emissions, and how can we better serve them?
\end{itemize}

\section{Methods \& Results}
We break the total environmental cost of the CV pipeline into two sub-components: (1) the cost of searching for an architecture and (2) the evaluation cost of an architecture over its lifetime. Within this work, the evaluation cost refers to CV researchers \& practitioners training a pre-established model. We attempt to quantify both types of costs and produce a quantitative metric for how our actions today are generating CO2 emissions.

\textbf{Searching Phase's CO2 emissions:} In order to determine the amount of CO2 emissions from CV-NAS searching phase, we used a Strubell et al. \cite{strubell-etal-2019-energy} inspired methodology. We begin by collecting the 45 major CV-NAS papers in the last three years (2018-2020) \cite{brock2018smash, ashok2018nn, ZhongYWSL18,CaiYZHY18, liu2018hierarchical, BenderKZVL18, LuoTQCL18, data_nips, fast_practical, xie2018snas, multinomial, bashivan2019teacher, Xiong_2019_ICCV, dong2019network, dong2019one, eff_fwd_nas, liu2018darts, cai2018proxylessnas, zhang2018graph, DongY19,  fbnetv1, renas, reg_ev_im, peng2020cream, shi_bridging_2020, chen2020anti, Hu2020TFNAS, chu2019fairdarts, dai2020data, he2020milenas, li2020sgas, wan2020fbnetv2, li2020block,fang2020densely, mei2019atomnas, Cai2020Once, wang2020attentivenas, luo2020neural, lu2021neural, chu2020noisy, chu2019scarlet, hundt2019sharpdarts, yu2020bignas, lu2020muxconv, Zoph_2018_CVPR} finding 157 models. For every model, we extract the Top-1 Accuracy, Parameters, FLOPS, GPU hours and GPU type. Using the GPU type, we found the total power draw for that specific GPU \cite{techpowerup_2021} denoted as $p_g$. If no GPU type was found, we assumed the most common GPU type, the Nvidia V100 was used. Often CPU type and DRAM weren't included in the papers, so we disregarded DRAM and assumed the Intel Core i7-10750H was used as the CPU power draw, denoted $p_c$. We then multiply the combined power draw by the GPU hours and by the Power Usage Effectiveness, 1.59 \cite{ascierto} to obtain a total power metric. We then denote this total power draw as $P_t$ in Watt-hours by:

\begin{equation}
    P_t \text{ [Wh]} = 1.59 \times \text{GPU hours} \times (p_g + p_c) 
    \label{eq:power}
\end{equation}

We then convert Watt-hours to CO2 emissions through the US Environmental Protection Agency's (EPA) kWh to CO2 measurement. This value was taken as a national weighted average of all CO2 emissions across the United States, giving 0.707 $\times 10^{-3}$ metric tonnes/kWh \cite{epa_2021}.  

\begin{equation}
    \text{CO2 [kg]} = P_t \times 0.707\times10^{-3}
    \label{eq:co2}
\end{equation}

\textbf{Evaluation Phase's CO2 emissions:}  As models get used over time, their FLOPS become a crucial factor in their computational cost. We can assess a models evaluation cost over time, as the number of times a model is being trained compared to the CO2-emissions. We denote a model's FLOPS as $f$ and quantify the Watt-to-FLOPS ratio as $\omega = \text{Watt}/f$. We then denote the GPU Watt-to-FLOPS as $\omega_g$ and the CPU's Watt-to-FLOPS as $\omega_c$. We obtain a model's power draw, $P_m$ over training by multiplying the model's flops, $f$ by the Watt-to-FLOPS then multiply the total GPU hours to train, seen below:

\begin{equation}
    P_m \text{ [Wh]}= f \times ( \omega_g + \omega_c) \times \text{GPU hours}
    \label{eq:flop_pow}
\end{equation}

Similar to \autoref{eq:co2} we can then compute the emitted CO2 by multiplying the power draw from a model by the EPA's Wh to CO2 measurement, CO2 = $P_m \cdot 0.707\times10^{-3} $.

\subsection{Results}\label{sec:results}
\textbf{Searching Phase:} Using Eq. \ref{eq:power} and Eq. \ref{eq:co2} we can quantify CV-NAS' searching phase's CO2 cost. These initial costs can then be amortized over time to observe how the searched architectures perform compared to the most popular hand-crafted networks like ResNet \cite{he2016deep}. With hand-crafted networks, the heuristics used are often derived over multiple trials and iterative training methods until satisfactory. These experiments are often never reported, making it difficult to quantify the hand-crafted network's CO2 cost. 



\begin{figure}[htp]
    \begin{center}
      \includegraphics[width=\linewidth]{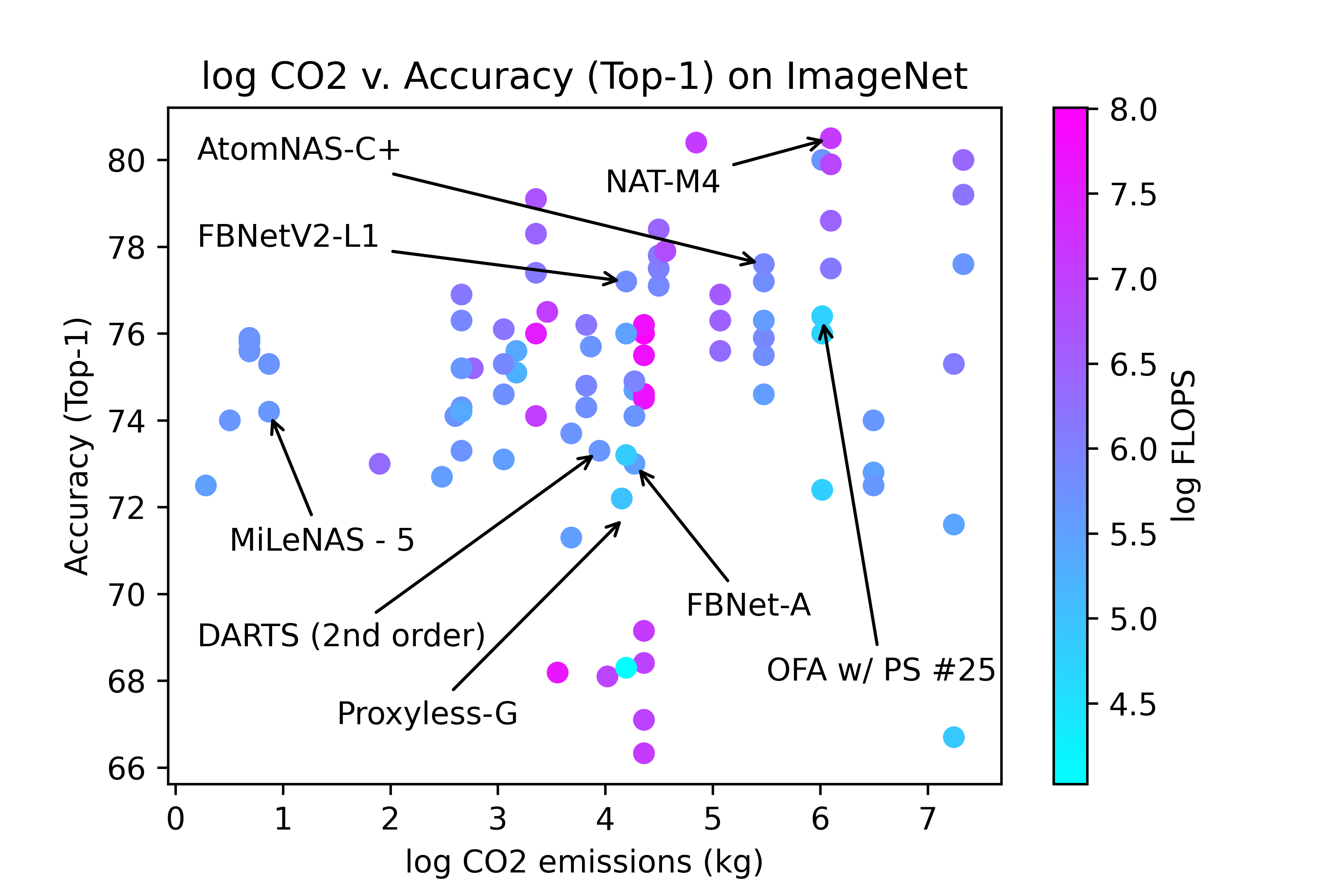}
    \end{center}
    \caption{CO2 emissions of CV-NAS optimized models compared to their initial CO2 searching cost.}
    \label{fig:imnet_co2}
\end{figure}

CV-NAS as an automated framework must have a trade-off between the top-1 accuracy compared to the CO2-emissions and relative FLOPS of a model. We highlight a few important models below in \autoref{fig:imnet_co2}, namely: NAT-M4 \cite{lu2021neural}, DARTS \cite{liu2018darts}, Once-For-All (OFA) \cite{Cai2020Once} and FB-Net A \cite{fbnetv1}. These models demonstrate how CV-NAS can trade off the initial CO2 cost for either accuracy or computational complexity in terms of FLOPS.  

In \autoref{fig:imnet_co2} we can see how the average FLOP models like MileNAS-5 have lower CO2 emissions but also average top-1 accuracy. Observe how OFA has lower computational complexity but also has a large initial CO2 cost, but doesn't beat the SOTA. Therefore, we can then trade-off the computational complexity to achieve SOTA by reducing our FLOP constraint, yielding NAT-M4. Low FLOP models like FBNet-A \cite{fbnetv1} and Proxyless-G \cite{cai2018proxylessnas} generally have average initial CO2 cost but only average accuracy.

\begin{figure}[htp]
    \begin{center}
      \includegraphics[width=\linewidth]{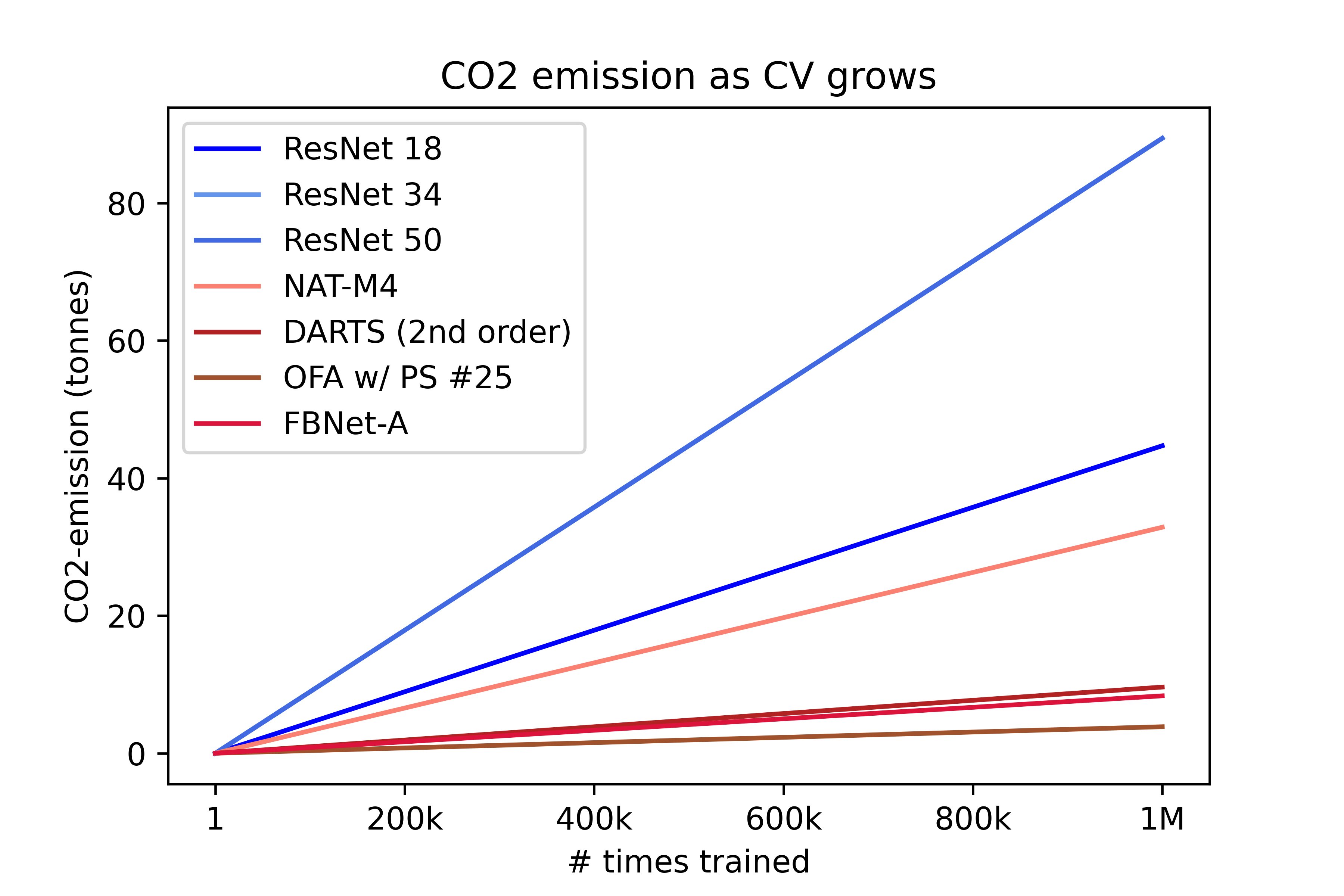}
    \end{center}
    \vspace{-0.4cm}
    \caption{CO2 emissions of CV-NAS compared to heuristically optimized networks per 1 million times trained on ImageNet. The initial searching cost is negligible compared to the enormous evaluation cost over time.}
    \label{fig:co2_v_times_trained}
\end{figure}

\textbf{Evaluation Phase:} When a new model is released and becomes SOTA, it quickly becomes popular and can amass thousands of citations. For example, the ResNet architecture \cite{he2016deep} published in CVPR 2016 has currently amassed over 73k citations \cite{goog_most_pop_arch} and is considered the standard for research. Assuming each paper ran their ResNet-backbone a modest 50 times to determine heuristic optimizations for their model, ResNet has been trained \textit{$\approx$3.6 million times}. Therefore, model adoption has a huge impact on the total computational burden required to train a model, and by consistently using sub-optimal networks, we are incurring a large CO2 cost from evaluation. These large CO2 costs can be seen in \autoref{fig:co2_v_times_trained}, which was calculated by assuming 250 epochs at 40 mins/epoch for ResNet and 60 mins/epoch for CV-NAS models. Then, we used Eq. \ref{eq:flop_pow} to derive a model's power draw and equivalent CO2 emission. The models chosen to compare to ResNet in \autoref{fig:co2_v_times_trained} were to demonstrate the trade-offs between CO2-Accuracy-FLOPS.

Furthermore, we can also compare how hand-crafted networks compare to each other over their lifetimes so far. We do so by multiplying Eq \ref{eq:flop_pow} by the number of citations and 50 times trained per citation yielding \autoref{fig:co2_hand_crafted} below.

\begin{figure}[H]
    \begin{center}
      \includegraphics[width=0.8\linewidth,height=5cm,keepaspectratio,]{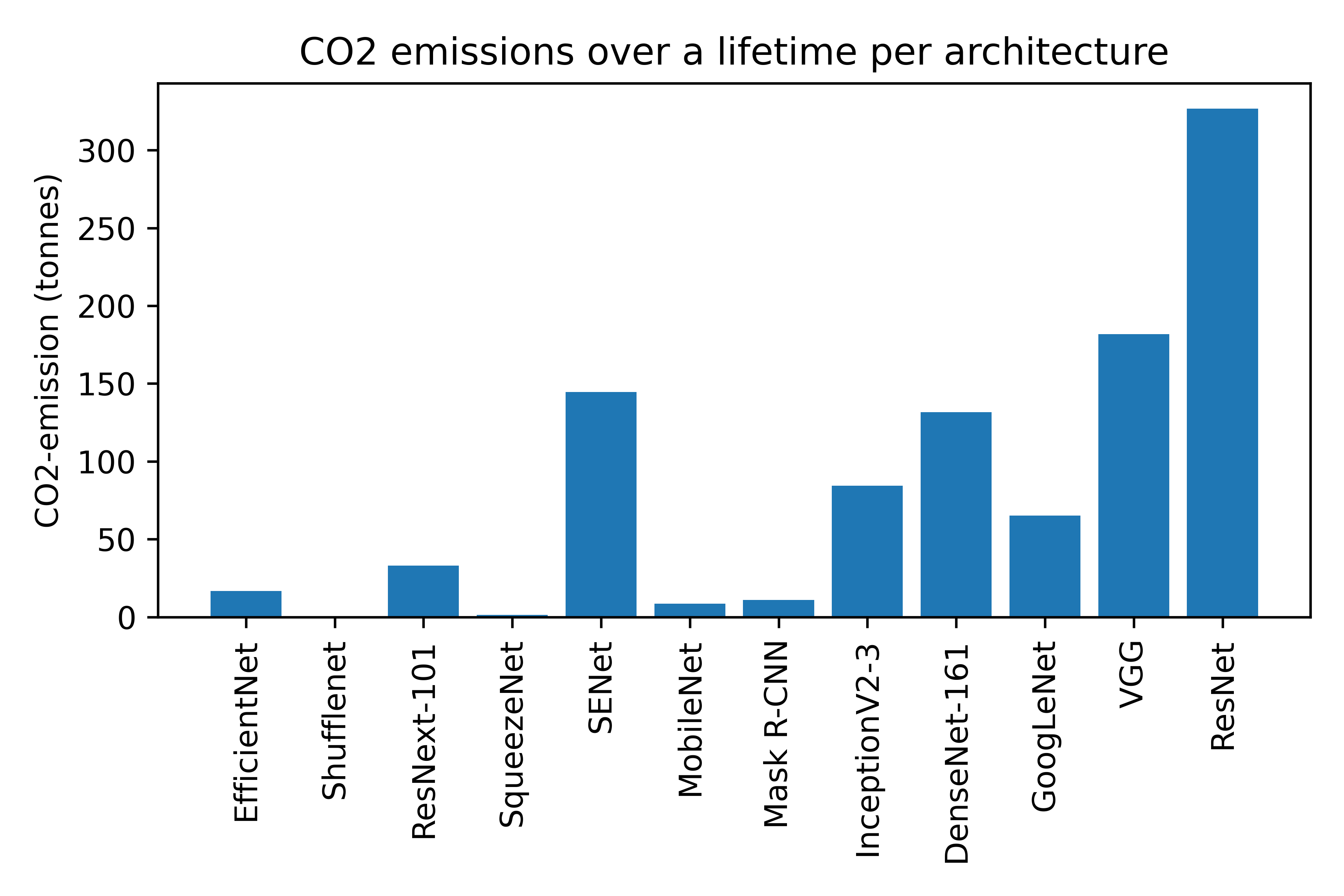}
    \end{center}
    \vspace{-0.5cm}
    \caption{CO2 emissions of hand-crafted models over their lifetimes so far. The common architectures are ordered from left-to-right as lowest-to-highest number of citations}
    \label{fig:co2_hand_crafted}
\end{figure}


\section{Discussion}
To quantify the total amount of CO2 emissions emitted, we isolated two components: (1) the initial searching phase and (2) the evaluation phase. By addressing how each phase of a model's life-cycle can be optimized to reduce CO2 emissions, we can lower the overall CO2-cost. We first identify how CO2 emissions tie into the global climate crisis, and identify marginalized groups who will be adversely affected by our actions. We then suggest some possible solutions to the CV-NAS and broader CV community. 

\textbf{The Current Climate Crisis:}
According to the Intergovernmental Panel on Climate Change (IPCC) \cite{ipcc_2019} we are currently encountering a climate crisis that must be capped at a 1.5$^{\circ}$C change. They highlight that while 1.5$^{\circ}$ C is the ideal goal, we are currently not on track to reach a limit of 1.5$^{\circ}$C. In 2016, we emitted 52 GtCO2 and by 2030 we will have 52-58 GtCO2, while we should be at 25-30 GtCO2 within the same timeframe. This crisis highlights how vital it is that we, within the CV-community enact change as any reduction in CO2 emissions can help curb the climate crisis. Additionally, the IPCC \cite{ipcc_2019} highlights the unprecedented level of cooperation required between inter-government and inter-personal to reduce CO2 output to desirable levels. 

The natural question underpinning this paper is the impact the CV community has on the CO2 emissions and if we can even make a substantive change to our environmental future. We believe that this question is valid, but mistakes the core idea behind curbing climate emissions, that everyone has a role to play, even the CV community.

\textbf{Stakeholders:} 
CO2 emissions are becoming a severe issue within society with considerable affect on the future, as demonstrated by the IPCC report. While everyone today and in the future is affected by the climate crisis, we scope this discussion to a rarely discussed marginalized group: our future generations. In identifying this group we discover subgroups who will have concrete impacts due to our CO2 emissions. Two such subgroups that will be impacted are those in the coastal and equatorial regions. In equatorial Africa, between 75 and 250 million people will be experiencing increased water-stress and rain-fed agriculture will be reduced by up to 50\%. \cite{ipcc_2019} Those in coastal regions will experience between 1 to 8 feet of sea level increase, engulfing their homes and livelihoods. These subgroups of future generations require immediate changes to our current CO2 emissions to minimize the effects of the climate crisis on them, and our role within as researchers within the CV community, should be to attempt to curb our emissions where possible. 

\textbf{Ethical considerations: }
Within the Ethical AI community, we’ve adopted \cite{jobin_ienca_vayena_2019, ai4people, whittlestone_2019} Principalism's \cite{beever_brightman_2015} four pillars: Respect for autonomy, Beneficence, Non-maleficence and Justice to analyze ethical concerns that have risen from computer-vision based dilemmas such as dataset bias and identifying marginalized communities. Through principalism, we can also analyze how stakeholders are unintentionally being affected by our actions today. In particular how we are overlooking all four pillars for the future generations, in effect showing that the ethical AI frameworks within CV are ill-enforced. Firstly, by emitting this level of atmospheric CO2, we overlook the future generation's capacity to be self-determining by forcing them to cope with incredible amounts of agricultural stress and increased risk for extreme weather events. Secondly, we are currently disregarding beneficence by not acting in the benefit of future generations. Our unintended actions aren’t only non-benevolent but also maleficent. By causing harm to future generations despite the warnings, we are inadvertently acting in direct contraposition to the ethical framework we’ve established. Finally, by emitting CO2 today that will have severe negative implications for future generations, we’re taking on the benefit while shifting the environmental cost to the future generations thus violating Justice. As such, we can see that despite establishing principalism as our working ethical framework, our unforeseen actions in emitting CO2 are going against those values.

\subsection{Future Design Considerations}\label{sec:altsoln}
While the AI community has adopted principalism as a foundational tenet of ethical AI, the four pillars don't necessarily reflect current societal and environmental considerations. As researchers, we have an obligation to maintain ethical considerations for marginalized groups. Therefore, we propose a \textit{fifth pillar} for ethical AI, \textit{enforcement}. Enforcement is the idea that our obligation to the marginalized groups should be included within our research, and these obligations should permeate our work. By holding ourselves accountable, we can act responsibly and attempt to minimize the global climate crisis. 

\textbf{CV-NAS community:} Within the CV-NAS community we can adopt \textit{enforcement} by incorporating CO2 considerations, providing models that not only minimize CO2 during search, but also ones that minimize over lifetime of the model. During initial searching, these models trade off accuracy at any computational and environmental cost. 


By incorporating both types of CO2 cost into the optimization, newer models can balance CO2-emissions to the generated architecture's optimization. We further highlight that while CV-NAS models have an initial searching cost (Fig. \ref{fig:imnet_co2}) it is often dwarfed by the life-time evaluation cost (Fig. \ref{fig:co2_v_times_trained}). Therefore, by optimizing on CO2 for searching, but prioritizing low FLOP models, CV-NAS can have a considerable impact on CO2 emissions as a whole. In particular we note how the 3 branches of CV-NAS can better optimized: (a) \textit{NAS-RL}, building on hardware-aware NAS to include a CO2/FLOP objective, (b) \textit{Gradient-based}, including a CO2/FLOP constraint on the chosen-operations level of the bi-level \cite{liu2018darts} and (c) \textit{Evolution}, adding a CO2/FLOP regularization parameter to the fitness function.

CV-NAS, like most of deep-learning, suffers from a lack of explainability. Often, the models are generated through black-box approaches, which may optimize for different criteria. We suggest that further research be done into CV-NAS explainability, as a glass-box approach to model optimization would allow the community to directly manipulate optimization parameters, meaning a greater focus on FLOP reduction which curbs long-term CO2 emissions. Furthermore, by incorporating FLOPS as a constraint we can also reduce the total cost over a lifetime seen in \autoref{fig:co2_v_times_trained}. We can \textit{enforce} optimizing on FLOPS by refusing to use high FLOP networks, thereby communally penalizing researchers publishing these types of models. Optimizing on FLOPS has the added bonus of reduced evaluation computational power making them ideal for mobile, edge and low-power devices. 



\begin{table}[H]
\centering
\begin{tabular}{c|c|c|c}
                   & \textbf{CO2 (tons)} & \textbf{\begin{tabular}[c]{@{}c@{}}Cars \\ Driven\end{tabular}} & \textbf{\begin{tabular}[c]{@{}c@{}}Homes \\ Powered\end{tabular}} \\ \hline
\textit{ResNet} \cite{he2016deep}   & 326.6               & 70.6                                                            & 55.3                                                              \\ \hline
\textit{VGG} \cite{simonyan2014very}& 181.7               & 39.8                                                            & 30.8                                                              \\ \hline
\textit{GoogLeNet} \cite{szegedy2015going} & 65.1                & 14.1                                                            & 11                                                               
\end{tabular}
\caption{To contextualize CO2 costs from the 3 most popular hand-crafted models (numeric values from Fig. \ref{fig:co2_hand_crafted}), we contrast with cars driven \& homes powered over 1 year. \cite{epa_calc_2018}.  \label{tab:equiv_co2} }
\end{table}
\vspace{-4mm}
\textbf{Broader CV community:}  In \autoref{fig:co2_v_times_trained} and \autoref{tab:equiv_co2} we can see that over time, choosing a sub-optimal model to conduct experiments has a severe and quantitative impact on the total CO2 produced. In line with \textit{enforcement} we propose that the broader CV community adopt searched architectures from CV-NAS as baseline models in order to curb the amortized CO2 cost. If reviewers question why high FLOP models, like ResNet, was used as the optimized model over a lower FLOP cost model, our community can \textit{enforce} the ethical AI principles. By adopting lower FLOP models such as FBNet-A or OFA, we can reduce the CO2 cost over a lifetime, allowing the broader CV community to also curb CO2 emissions. 

One of the main reasons behind high-FLOP network development is image-scaling. Popular datasets like ImageNet \cite{imagenet_cvpr09} dominate research as the standard for generalization, but it come with high-FLOPS. Therefore, in order to curb the life-time evaluation CO2 cost, we encourage the CV-community to consider downscaled benchmarks for training \& development such as Tiny-ImageNet \cite{le2015tiny}.

\section{Conclusion}
Through this paper, we investigated a vitally important problem that is threatening our lives: Climate Change, and how the CV-NAS and broader CV community are contributing to the climate crisis. We then investigated the two main components behind CV CO2 emissions, (1) the initial architecture creation and (2) the evaluation of the network over its lifetime. We then provide a novel analysis of the CO2 emissions from CV-NAS and the CV-community, demonstrating that emissions are non-negligible. We then discuss which communities are at most risk and discuss how our community is inadvertently disregarding the Ethical AI principles. Therefore, we proposed a set of recommendations for the CV-NAS and broader CV community in order to curb the climate crisis. Further work can be done in this domain, linking environmental racism to CO2 output and the unintended consequences of the climate crisis. 

{\small
\bibliographystyle{ieee_fullname}
\bibliography{egbib}

\begin{thebibliography}{10}\itemsep=-1pt

\bibitem{ipcc_2019}
Climate change and land: an ipcc special report on climate change,
  desertification, land degradation, sustainable land management, food
  security, and greenhouse gas fluxes in terrestrial ecosystems.
\newblock {\em https://www.ipcc.ch/srccl}.

\bibitem{epa_calc_2018}
Epa greenhouse gases calculator, Oct 2018.

\bibitem{goog_most_pop_arch}
Nature index: Most influential papers research citations, Jul 2020.

\bibitem{epa_2021}
Epa greenhouse gases equivalencies references, Feb 2021.

\bibitem{techpowerup_2021}
Techpowerup, recent database updates, Feb 2021.

\bibitem{ascierto}
Rhonda Ascierto.
\newblock Uptime institute global data center survey 2020.
\newblock {\em Uptime Institute}.

\bibitem{ashok2018nn}
Anubhav Ashok, Nicholas Rhinehart, Fares Beainy, and Kris~M. Kitani.
\newblock N2n learning: Network to network compression via policy gradient
  reinforcement learning.
\newblock In {\em International Conference on Learning Representations}, 2018.

\bibitem{bashivan2019teacher}
Pouya Bashivan, Mark Tensen, and James~J DiCarlo.
\newblock Teacher guided architecture search, 2019.

\bibitem{beever_brightman_2015}
Jonathan Beever and Andrew~O. Brightman.
\newblock Reflexive principlism as an effective approach for developing ethical
  reasoning in engineering.
\newblock {\em Science and Engineering Ethics}, 22(1):275–291, 2015.

\bibitem{BenderKZVL18}
Gabriel Bender, Pieter-Jan Kindermans, Barret Zoph, Vijay Vasudevan, and
  Quoc~V. Le.
\newblock Understanding and simplifying one-shot architecture search.
\newblock In {\em ICML}, pages 549--558, 2018.

\bibitem{brock2018smash}
Andrew Brock, Theo Lim, J.M. Ritchie, and Nick Weston.
\newblock {SMASH}: One-shot model architecture search through hypernetworks.
\newblock In {\em International Conference on Learning Representations}, 2018.

\bibitem{Cai2020Once}
Han Cai, Chuang Gan, Tianzhe Wang, Zhekai Zhang, and Song Han.
\newblock Once-for-all: Train one network and specialize it for efficient
  deployment.
\newblock In {\em International Conference on Learning Representations}, 2020.

\bibitem{CaiYZHY18}
Han Cai, Jiacheng Yang, Weinan Zhang, Song Han, and Yong Yu.
\newblock Path-level network transformation for efficient architecture search.
\newblock In {\em ICML}, pages 677--686, 2018.

\bibitem{cai2018proxylessnas}
Han Cai, Ligeng Zhu, and Song Han.
\newblock Proxyless{NAS}: Direct neural architecture search on target task and
  hardware.
\newblock In {\em International Conference on Learning Representations}, 2019.

\bibitem{data_nips}
Jianlong Chang, xinbang zhang, Yiwen Guo, GAOFENG MENG, SHIMING XIANG, and
  Chunhong Pan.
\newblock Data: Differentiable architecture approximation.
\newblock In H. Wallach, H. Larochelle, A. Beygelzimer, F. d\textquotesingle
  Alch\'{e}-Buc, E. Fox, and R. Garnett, editors, {\em Advances in Neural
  Information Processing Systems}, volume~32. Curran Associates, Inc., 2019.

\bibitem{chen2020anti}
Hanlin Chen, Baochang Zhang, Song Xue, Xuan Gong, Hong Liu, Rongrong Ji, and
  David Doermann.
\newblock Anti-bandit neural architecture search for model defense.
\newblock In {\em European Conference on Computer Vision}, pages 70--85.
  Springer, 2020.

\bibitem{renas}
Yukang Chen, Gaofeng Meng, Qian Zhang, Shiming Xiang, Chang Huang, Lisen Mu,
  and Xinggang Wang.
\newblock Renas: Reinforced evolutionary neural architecture search.
\newblock In {\em CVPR}, pages 4787--4796, 2019.

\bibitem{chu2019scarlet}
Xiangxiang Chu, Bo Zhang, Jixiang Li, Qingyuan Li, and Ruijun Xu.
\newblock Scarlet-nas: bridging the gap between stability and scalability in
  weight-sharing neural architecture search.
\newblock {\em arXiv preprint arXiv:1908.06022}, 2019.

\bibitem{chu2020noisy}
Xiangxiang Chu, Bo Zhang, and Xudong Li.
\newblock Noisy differentiable architecture search.
\newblock {\em arXiv preprint arXiv:2005.03566}, 2020.

\bibitem{chu2019fairdarts}
Xiangxiang Chu, Tianbao Zhou, Bo Zhang, and Jixiang Li.
\newblock {Fair DARTS: Eliminating Unfair Advantages in Differentiable
  Architecture Search}.
\newblock In {\em 16th Europoean Conference On Computer Vision}, 2020.

\bibitem{fast_practical}
J. {Cui}, P. {Chen}, R. {Li}, S. {Liu}, X. {Shen}, and J. {Jia}.
\newblock Fast and practical neural architecture search.
\newblock In {\em 2019 IEEE/CVF International Conference on Computer Vision
  (ICCV)}, pages 6508--6517, 2019.

\bibitem{dai2020data}
Xiyang Dai, Dongdong Chen, Mengchen Liu, Yinpeng Chen, and Lu Yuan.
\newblock Da-nas: Data adapted pruning for efficient neural architecture
  search.
\newblock {\em arXiv preprint arXiv:2003.12563}, 2020.

\bibitem{imagenet_cvpr09}
J. Deng, W. Dong, R. Socher, L.-J. Li, K. Li, and L. Fei-Fei.
\newblock {ImageNet: A Large-Scale Hierarchical Image Database}.
\newblock In {\em CVPR09}, 2009.

\bibitem{dong2019network}
Xuanyi Dong and Yi Yang.
\newblock Network pruning via transformable architecture search.
\newblock {\em arXiv preprint arXiv:1905.09717}, 2019.

\bibitem{dong2019one}
Xuanyi Dong and Yi Yang.
\newblock One-shot neural architecture search via self-evaluated template
  network.
\newblock In {\em Proceedings of the IEEE/CVF International Conference on
  Computer Vision}, pages 3681--3690, 2019.

\bibitem{DongY19}
Xuanyi Dong and Yi Yang.
\newblock Searching for a robust neural architecture in four gpu hours.
\newblock In {\em CVPR}, pages 1761--1770, 2019.

\bibitem{fang2020densely}
Jiemin Fang, Yuzhu Sun, Qian Zhang, Yuan Li, Wenyu Liu, and Xinggang Wang.
\newblock Densely connected search space for more flexible neural architecture
  search.
\newblock In {\em Proceedings of the IEEE/CVF Conference on Computer Vision and
  Pattern Recognition}, pages 10628--10637, 2020.

\bibitem{ai4people}
Luciano Floridi, Josh Cowls, Monica Beltrametti, Raja Chatila, Patrice
  Chazerand, Virginia Dignum, Christoph Luetge, Robert Madelin, Ugo Pagallo,
  Francesca Rossi, and et al.
\newblock Ai4people—an ethical framework for a good ai society:
  Opportunities, risks, principles, and recommendations.
\newblock {\em Minds and Machines}, 28(4):689–707, 2018.

\bibitem{he2020milenas}
Chaoyang He, Haishan Ye, Li Shen, and Tong Zhang.
\newblock Milenas: Efficient neural architecture search via mixed-level
  reformulation.
\newblock In {\em Proceedings of the IEEE/CVF Conference on Computer Vision and
  Pattern Recognition}, pages 11993--12002, 2020.

\bibitem{he2016deep}
Kaiming He, Xiangyu Zhang, Shaoqing Ren, and Jian Sun.
\newblock Deep residual learning for image recognition.
\newblock In {\em Proceedings of the IEEE conference on computer vision and
  pattern recognition}, pages 770--778, 2016.

\bibitem{eff_fwd_nas}
Hanzhang Hu, John Langford, Rich Caruana, Saurajit Mukherjee, Eric Horvitz, and
  Debadeepta Dey.
\newblock Efficient forward architecture search.
\newblock {\em CoRR}, abs/1905.13360, 2019.

\bibitem{hundt2019sharpdarts}
Andrew Hundt, Varun Jain, and Gregory~D Hager.
\newblock sharpdarts: Faster and more accurate differentiable architecture
  search.
\newblock {\em arXiv preprint arXiv:1903.09900}, 2019.

\bibitem{jobin_ienca_vayena_2019}
Anna Jobin, Marcello Ienca, and Effy Vayena.
\newblock The global landscape of ai ethics guidelines.
\newblock {\em Nature Machine Intelligence}, 1(9):389–399, 2019.

\bibitem{le2015tiny}
Ya Le and Xuan Yang.
\newblock Tiny imagenet visual recognition challenge.
\newblock {\em CS 231N}, 7:7, 2015.

\bibitem{li2020block}
Changlin Li, Jiefeng Peng, Liuchun Yuan, Guangrun Wang, Xiaodan Liang, Liang
  Lin, and Xiaojun Chang.
\newblock Block-wisely supervised neural architecture search with knowledge
  distillation.
\newblock In {\em Proceedings of the IEEE/CVF Conference on Computer Vision and
  Pattern Recognition}, pages 1989--1998, 2020.

\bibitem{li2020sgas}
Guohao Li, Guocheng Qian, Itzel~C Delgadillo, Matthias Muller, Ali Thabet, and
  Bernard Ghanem.
\newblock Sgas: Sequential greedy architecture search.
\newblock In {\em Proceedings of the IEEE/CVF Conference on Computer Vision and
  Pattern Recognition}, pages 1620--1630, 2020.

\bibitem{liu2018hierarchical}
Hanxiao Liu, Karen Simonyan, Oriol Vinyals, Chrisantha Fernando, and Koray
  Kavukcuoglu.
\newblock Hierarchical representations for efficient architecture search.
\newblock In {\em International Conference on Learning Representations}, 2018.

\bibitem{liu2018darts}
Hanxiao Liu, Karen Simonyan, and Yiming Yang.
\newblock {DARTS}: Differentiable architecture search.
\newblock In {\em International Conference on Learning Representations}, 2019.

\bibitem{lu2020muxconv}
Zhichao Lu, Kalyanmoy Deb, and Vishnu~Naresh Boddeti.
\newblock Muxconv: Information multiplexing in convolutional neural networks.
\newblock In {\em Proceedings of the IEEE/CVF Conference on Computer Vision and
  Pattern Recognition}, pages 12044--12053, 2020.

\bibitem{lu2021neural}
Zhichao Lu, Gautam Sreekumar, Erik Goodman, Wolfgang Banzhaf, Kalyanmoy Deb,
  and Vishnu~Naresh Boddeti.
\newblock Neural architecture transfer.
\newblock {\em IEEE Transactions on Pattern Analysis and Machine Intelligence},
  2021.

\bibitem{luo2020neural}
Renqian Luo, Xu Tan, Rui Wang, Tao Qin, Enhong Chen, and Tie-Yan Liu.
\newblock Neural architecture search with gbdt.
\newblock {\em arXiv preprint arXiv:2007.04785}, 2020.

\bibitem{LuoTQCL18}
Renqian Luo, Fei Tian, Tao Qin, Enhong Chen, and Tie-Yan Liu.
\newblock Neural architecture optimization.
\newblock In {\em NeurIPS}, pages 7827--7838, 2018.

\bibitem{mei2019atomnas}
Jieru Mei, Yingwei Li, Xiaochen Lian, Xiaojie Jin, Linjie Yang, Alan Yuille,
  and Jianchao Yang.
\newblock Atomnas: Fine-grained end-to-end neural architecture search.
\newblock {\em arXiv preprint arXiv:1912.09640}, 2019.

\bibitem{peng2020cream}
Houwen Peng, Hao Du, Hongyuan Yu, Qi Li, Jing Liao, and Jianlong Fu.
\newblock Cream of the crop: Distilling prioritized paths for one-shot neural
  architecture search.
\newblock {\em Advances in Neural Information Processing Systems}, 33, 2020.

\bibitem{reg_ev_im}
Esteban Real, Alok Aggarwal, Yanping Huang, and Quoc~V. Le.
\newblock Regularized evolution for image classifier architecture search.
\newblock In {\em AAAI}, pages 4780--4789, 2019.

\bibitem{shi_bridging_2020}
Han Shi, Renjie Pi, Hang Xu, Zhenguo Li, James Kwok, and Tong Zhang.
\newblock Bridging the gap between sample-based and one-shot neural
  architecture search with bonas.
\newblock In H. Larochelle, M. Ranzato, R. Hadsell, M.~F. Balcan, and H. Lin,
  editors, {\em Advances in Neural Information Processing Systems}, volume~33,
  pages 1808--1819. Curran Associates, Inc., 2020.

\bibitem{simonyan2014very}
Karen Simonyan and Andrew Zisserman.
\newblock Very deep convolutional networks for large-scale image recognition.
\newblock {\em arXiv preprint arXiv:1409.1556}, 2014.

\bibitem{strubell-etal-2019-energy}
Emma Strubell, Ananya Ganesh, and Andrew McCallum.
\newblock Energy and policy considerations for deep learning in {NLP}.
\newblock In {\em Proceedings of the 57th Annual Meeting of the Association for
  Computational Linguistics}, pages 3645--3650, Florence, Italy, July 2019.
  Association for Computational Linguistics.

\bibitem{szegedy2015going}
Christian Szegedy, Wei Liu, Yangqing Jia, Pierre Sermanet, Scott Reed, Dragomir
  Anguelov, Dumitru Erhan, Vincent Vanhoucke, and Andrew Rabinovich.
\newblock Going deeper with convolutions.
\newblock In {\em Proceedings of the IEEE conference on computer vision and
  pattern recognition}, pages 1--9, 2015.

\bibitem{wan2020fbnetv2}
Alvin Wan, Xiaoliang Dai, Peizhao Zhang, Zijian He, Yuandong Tian, Saining Xie,
  Bichen Wu, Matthew Yu, Tao Xu, Kan Chen, et~al.
\newblock Fbnetv2: Differentiable neural architecture search for spatial and
  channel dimensions.
\newblock In {\em Proceedings of the IEEE/CVF Conference on Computer Vision and
  Pattern Recognition}, pages 12965--12974, 2020.

\bibitem{wang2020attentivenas}
Dilin Wang, Meng Li, Chengyue Gong, and Vikas Chandra.
\newblock Attentivenas: Improving neural architecture search via attentive
  sampling.
\newblock {\em arXiv preprint arXiv:2011.09011}, 2020.

\bibitem{whittlestone_2019}
Jess Whittlestone, Rune Nyrup, Anna Alexandrova, and Stephen Cave.
\newblock The role and limits of principles in ai ethics.
\newblock {\em Proceedings of the 2019 AAAI/ACM Conference on AI, Ethics, and
  Society}, 2019.

\bibitem{fbnetv1}
Bichen Wu, Xiaoliang Dai, Peizhao Zhang, Yanghan Wang, Fei Sun, Yiming Wu,
  Yuandong Tian, Peter Vajda, Yangqing Jia, and Kurt Keutzer.
\newblock Fbnet: Hardware-aware efficient convnet design via differentiable
  neural architecture search.
\newblock In {\em CVPR}, pages 10734--10742, 2019.

\bibitem{wu2015deep}
Ren Wu, Shengen Yan, Yi Shan, Qingqing Dang, and Gang Sun.
\newblock Deep image: Scaling up image recognition, 2015.

\bibitem{xie2018snas}
Sirui Xie, Hehui Zheng, Chunxiao Liu, and Liang Lin.
\newblock {SNAS}: stochastic neural architecture search.
\newblock In {\em International Conference on Learning Representations}, 2019.

\bibitem{Xiong_2019_ICCV}
Yunyang Xiong, Ronak Mehta, and Vikas Singh.
\newblock Resource constrained neural network architecture search: Will a
  submodularity assumption help?
\newblock In {\em Proceedings of the IEEE/CVF International Conference on
  Computer Vision (ICCV)}, October 2019.

\bibitem{Hu2020TFNAS}
Xiang~Wu Yibo~Hu and Ran He.
\newblock Tf-nas: Rethinking three search freedoms of latency-constrained
  differentiable neural architecture search.
\newblock In {\em Proc. Eur. Conf. Computer Vision (ECCV)}, 2020.

\bibitem{yu2020bignas}
Jiahui Yu, Pengchong Jin, Hanxiao Liu, Gabriel Bender, Pieter-Jan Kindermans,
  Mingxing Tan, Thomas Huang, Xiaodan Song, Ruoming Pang, and Quoc Le.
\newblock Bignas: Scaling up neural architecture search with big single-stage
  models.
\newblock In {\em European Conference on Computer Vision}, pages 702--717.
  Springer, 2020.

\bibitem{zhang2018graph}
Chris Zhang, Mengye Ren, and Raquel Urtasun.
\newblock Graph hypernetworks for neural architecture search.
\newblock In {\em International Conference on Learning Representations}, 2019.

\bibitem{multinomial}
Xiawu Zheng, Rongrong Ji, Lang Tang, Baochang Zhang, Jianzhuang Liu, and Qi
  Tian.
\newblock Multinomial distribution learning for effective neural architecture
  search.
\newblock {\em CoRR}, abs/1905.07529, 2019.

\bibitem{ZhongYWSL18}
Zhao Zhong, Junjie Yan, Wei Wu, Jing Shao, and Cheng-Lin Liu.
\newblock Practical block-wise neural network architecture generation.
\newblock In {\em CVPR}, pages 2423--2432, 2018.

\bibitem{Zoph_2018_CVPR}
Barret Zoph, Vijay Vasudevan, Jonathon Shlens, and Quoc~V. Le.
\newblock Learning transferable architectures for scalable image recognition.
\newblock In {\em Proceedings of the IEEE Conference on Computer Vision and
  Pattern Recognition (CVPR)}, June 2018.

\end{thebibliography}
}

\end{document}